\documentclass[conference]{IEEEtran}
\IEEEoverridecommandlockouts
\usepackage{cite}
\usepackage{amsmath,amssymb,amsfonts}
\usepackage{algorithmic}
\usepackage{graphicx}
\usepackage{textcomp}

\usepackage{color,soul}
\usepackage{algorithm}
\usepackage[table,xcdraw,usenames,dvipsnames]{xcolor}
\usepackage{subcaption}
\usepackage{enumitem}
\usepackage{url}
\def\BibTeX{{\rm B\kern-.05em{\sc i\kern-.025em b}\kern-.08em
    T\kern-.1667em\lower.7ex\hbox{E}\kern-.125emX}}
\begin{document}

\title{Failure-Sentient Composition For Swarm-Based Drone Services}

\author{\IEEEauthorblockN{Balsam Alkouz}
\IEEEauthorblockA{\textit{School of Computer Science} \\
\textit{The University of Sydney}\\
Sydney, Australia \\
balsam.alkouz@sydney.edu.au}
\and
\IEEEauthorblockN{Athman Bouguettaya}
\IEEEauthorblockA{\textit{School of Computer Science} \\
\textit{The University of Sydney}\\
Sydney, Australia \\
athman.bouguettaya@sydney.edu.au}
\and
\IEEEauthorblockN{Abdallah Lakhdari}
\IEEEauthorblockA{\textit{School of Computer Science} \\
\textit{The University of Sydney}\\
Sydney, Australia \\
abdallah.lakhdari@sydney.edu.au}
}

\maketitle
\thispagestyle{plain}
\pagestyle{plain}

\begin{abstract}
We propose a novel failure-sentient framework for swarm-based drone delivery services. The framework ensures that those drones that experience a noticeable degradation in their performance (called soft failure) and which are part of a swarm, do not disrupt the successful delivery of packages to a consumer. The framework composes a weighted continual federated learning prediction module to accurately predict the time of failures of individual drones and uptime after failures. These predictions are used to determine the severity of failures at both the drone and swarm levels. We propose a speed-based heuristic algorithm with lookahead optimization to generate an optimal set of services considering failures. Experimental results on real datasets prove the efficiency of our proposed approach in terms of prediction accuracy, delivery times, and execution times.
\end{abstract}


\section{Introduction}

Drone technology continues to evolve and prove itself useful in an ever-growing list of applications \cite{mueller2017drones}. As a logical progression, drone swarm technology represents the next major leap forward for drones. Drone swarm services are used in search and rescue \cite{drew2021multi}, sky shows \cite{waibel2017drone}, fire fighting \cite{innocente2019self}, and delivery \cite{alkouz2022flight}. Our primary focus is on the use of drone swarms in the delivery of packages. The majority of delivery drones in development today are designed to carry small payloads \cite{otto2018optimization}\cite{shahzaad2023optimizing}. Sending heavier packages would normally necessitate the use of a large drone. However, larger drones are being refused by regulators and the general public due to concerns about their safety, noise levels, and cost \cite{alkouz2021reinforcement}. As an alternative, swarm-based drone delivery services promise to deliver multiple/heavier packages to a destination simultaneously \cite{alkouz2020swarm}.

Swarm-based delivery services, also known as Swarm-based Drone-as-a-Service (SDaaS), promise the effective delivery of scalable orders. Several solutions have been presented to efficiently deliver packages from a source to a destination node in a skyway network utilising a swarm of drones \cite{alkouz2020formation} \cite{alkouz2022flight}. A skyway network is made up of charging stations installed on building rooftops \cite{lee2022autonomous}. Typically, a swarm would navigate the network from the source to the destination node. However, existing solutions focus on swarm routing and allocation, assuming a deterministic environment with no failures \cite{alkouz2021provider}. \looseness=-1

Failures in drones may occur as a result of environmental uncertainties such as weather conditions and the inherent specifications of drones \cite{torabbeigi2021optimization}. A drone, like any other technology, has a Mean Time To Failure (MTTF). MTTF is a maintenance metric that measures the average amount of time an asset operates before it fails. Failure in drone delivery is caused by a number of factors \cite{shahzaad2021robust}. This comprises the history of carried packages weights, the weather conditions encountered, and the distance traveled. Failures in drone swarms are often more severe than failures in single drone deliveries \cite{chung2018survey}. This research aims to reduce the impact of failures in drone swarms.

A \textit{delivery failure} is generally defined as an expected {\em late} arrival of packages to a consumer at the destination \cite{shahzaad2021robust}. A delivery failure is generally caused by a \textit{drone failure}. Therefore delivery failures are defined as resulting from drone failures. We identify two type of drone failures: \textit{hard} failures and \textit{soft} failures. A hard failure occurs when a drone crashes or ceases to operate \cite{petritoli2018reliability}. A soft failure, on the other hand, is a degradation in a drone's performance. This paper focuses on \textit{soft failures}, in which performance may decline at some time but the drone may recover after a while. Because the drones can repair themselves on reset, the Mean Time Between Failures (MTBF) reflects this idea \cite{petritoli2018reliability}. We focus on soft failure that manifest as increased battery usage \cite{magsino2020achieving}. A swarm failure is more severe than a single drone failure since the failure of a subset of the swarm affects the other drones. We assume a swarm is atomic, i.e. the drones stick together from the source to the destination \cite{akram2017security}. A failure of a subset of a swarm is referred to as a \textit{partial failure}. Because the swarm members are atomic, if a partial failure happens, the rest of the swarm members will need to accommodate.


The objective of this work is to reduce the impact of soft failures on the swarm and the delivery mission. A soft failures should have a negligible effect on delivery,  leaving the consumer with little to no noticeable change in anticipated delivery time \cite{torabbeigi2018drone}\cite{alkouz2022density}. The challenge is that MTTF and MTBF only reflect the average time to failure. As a result, we don't know how many and which drones are likely to fail. Existing solutions deal with failures reactively, that is, after a failure happens \cite{bjerknes2013fault}. The issue with reactive solutions is that they are often too late and risky, resulting in hard failures. 
Therefore, we propose predicting soft failing drones and their failure times, and then proactively reroute the path considering slowing or speeding up the drones to accommodate failing drones. We summarize our main contribution as follows:
\begin{itemize}
    \item A novel failure-sentient SDaaS framework.
    \item An SDaaS failure prediction using weighted continual federated learning.
    \item A failure severity evaluation method at the drone and swarm levels.
    \item A speed-based heuristic algorithm for failure-sentient composition with lookaheads optimization.
\end{itemize}

\begin{figure}[htbp!]
\centering
\includegraphics[width=\linewidth]{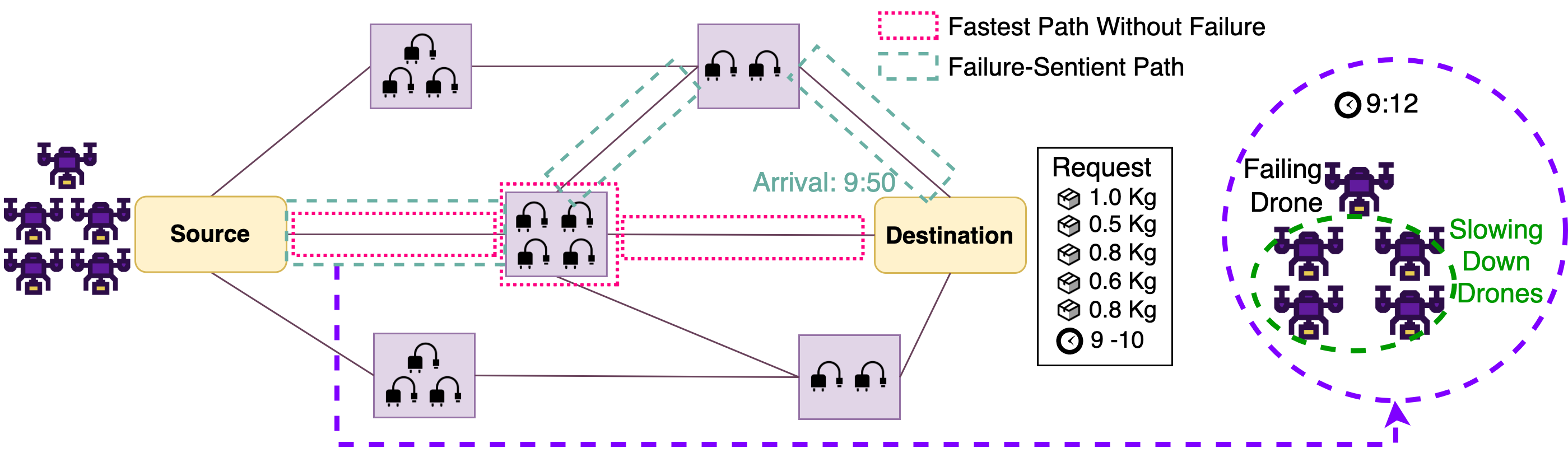}
\caption{Failure-Sentient Composition of Swarm Services in a Skyway Network} 
\label{motivating}
\end{figure}
\subsection{Motivating Scenario and Problem Statement}

Assume a medical facility requested that several pieces of medical equipment be delivered at the same time. Therefore, a swarm of drones is required to deliver these packages simultaneously. Let us also assume that the medical facility anticipates receiving the packages between 9:00 and 10:00 a.m. In an optimistic approach, assuming no failures, the swarm would proceed at high speeds along the fastest path, as illustrated in the dotted pink composed services in Fig. \ref{motivating}. We assume that an SDaaS service is a skyway segment serviced by a delivery swarm. This path would typically be optimal as the number of charging pads available would cater for more drones reducing the time of sequential charging.  Therefore, all drones would need to land to recharge to be able to continue the trip to the destination \cite{janszen2022constraint}.

Failure, however, is inevitable. Therefore, if a drone or a subset of the swarm fails at a certain point, the service composition algorithm should cater for the failing drones. In this work, we focus on challenges associated with soft failing drones, i.e. drone whose performance degrades. We assume a degradation would reflect in the form of extra battery consumption rate of a drone. If a drone fails, the impact on the swarm may be drastic due to the fact that they operate atomically, i.e. they stick together from the source to the destination. The swarm is assumed to be atomic because the arrival of packages simultaneously is bounded by a limited time window, e.g. 1-2 minutes. Therefore, we look into reducing the effect of soft failing drones on the whole swarm and in return on the delivery mission. 

\begin{figure}[htbp!]
\centering
\includegraphics[width=0.8\linewidth]{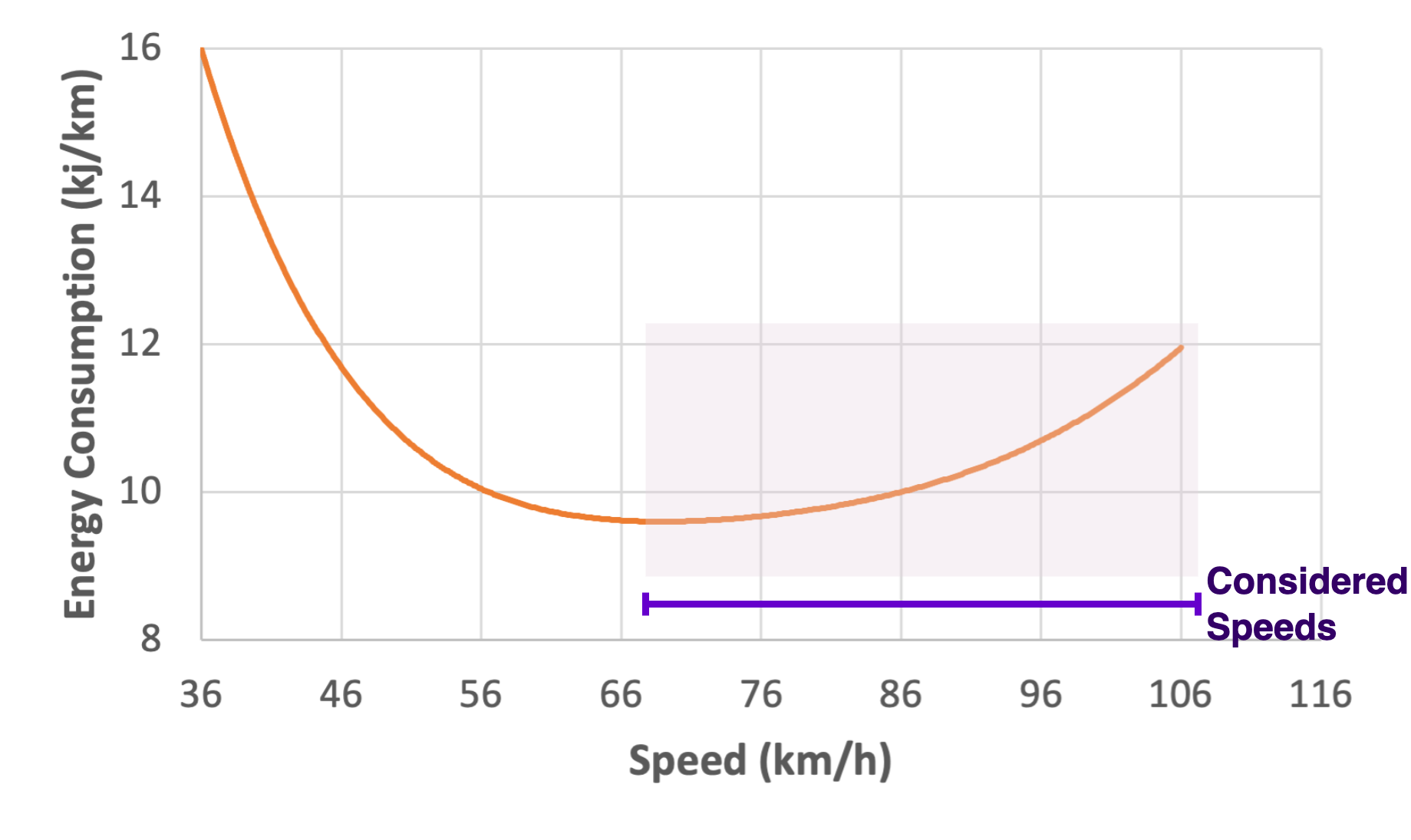}
\caption{Energy as a Function of Drone Speed \cite{dukkanci2021minimizing}} 
\label{speedEnergy}
\end{figure}

We propose to reduce the impact of failures on delivery missions by proactive measures. The preventive failure-sentient composition algorithm would compose different services depending on the degradation severity. The algorithm may command the healthy drones to slow down or speed up to cater for the needs of the soft failing drones. A drones speed affects its energy consumption. The energy consumption relation with speed is modeled as a convex function \cite{dukkanci2021minimizing}. For example, from speed 0 to 70km/h, a drones energy consumption reduces. A 70km/h speed would result in minimum consumption. Beyond 70km/h the energy consumption would keep increasing. We assume the base speed of the drones is around 105km/h, which is equivalent to the speed of the Alphabet Wing drone\footnote{\url{https://wing.com/en_au/how-it-works}}. We also assume that the speeds considered are in the range after the minimum as shown in Fig. \ref{speedEnergy}, such that decreasing the speed results in power consumption reduction and vise versa. In the example in Fig. \ref{motivating}, a drone is predicted to fail at 9:12 while the swarm is at the first segment. The degradation-based composition would command the rest of the drones to slow down to the lowest energy consuming speed, i.e. 70km/h, to cater for the need of the failing drone. As a result of slowing down, the drones consume less energy and do not need to stop at any node as it saves energy using the shorter composed path. This saves time and allows packages to be delivered sooner than if the original, longer route was used. Therefore, our proposed framework, should predict the soft failing drones time of failure and compute the failure severity. Then, the framework should compose the path taking into consideration soft failing drones. 

\section{Related Work}

A drone swarm is defined as a group of autonomous connected drones that work together to achieve common goals. 
Coordination and reactivity are critical because they constitute the essential difference between a true swarm and the mass deployment of single drones \cite{innocente2019self}. The majority of work in the literature refers to single drones deployed en masse as a drone swarm \cite{kuru2019analysis}. In contrast, we refer to a drone swarm as a group that communicates and travels together as a single entity \cite{alkouz2021service}. Current research on drone swarms for delivery focuses on the optimal service composition and allocation without taking into account uncertainties \cite{alkouz2022flight}\cite{alkouz2021reinforcement}. As a result, the purpose of this work is to fill a gap that is foreseeable and to suggest proactive approaches for anticipated failures.

In the literature, failure usually refers to the occurrence of a fault in a drone system \cite{miller2021survey}. Failures are typically categorised as system level or drone hardware/software level \cite{petritoli2017reliability}. The probability that a system will operate without failure is described as its reliability \cite{petritoli2018reliability}. A drone system's reliability can be calculated using mathematical models or measured and estimated using statistical parameters like Mean Time To Failure and Mean Time Between Failures \cite{petritoli2017reliability}.  Although swarm systems are often referred to as robust and scalable by default \cite{winfield2006safety}, the presence of failure in drone swarms is more drastic than a single drone failure. This is because the failing drones affect the rest of the swarm. As the swarm size increases the reliability of the swarm system drops \cite{bjerknes2013fault}.

Due to the detrimental effect of failures in swarm robotic systems, fault recovery mechanisms has been explored in the literature. For example, a fault recovery approach was proposed that uses reinforcement learning and self organising maps to select the most appropriate recovery strategy for any given scenario \cite{oladiran2019fault}. Moreover, to handle an agent failures during navigational process of a robots swarm, \cite{roy2020geometric} offered a system that can identify the faulty robots and thereby update the region’s information adaptively. Similarly, a trust-repairing method is proposed to restore performance and human trust in a swarm to an appropriate level by correcting undesired swarm behaviors \cite{liu2019trust}. As illustrated from the literature, swarm failures are typically addressed through corrective maintenance, i.e. dealing with failure after it occurs \cite{petritoli2018reliability}. We propose dealing with failure in a proactive manner, that is, before it occurs.


Despite the fact that addressing swarm failures has many dimensions, research into drone swarm failure is still in its infancy. Majority of work focuses on network link failures between drones in a swarm and the lead drone \cite{chen2020toward}\cite{chen2020achieving}. To the best of our knowledge, no previous work has been done on drone swarm failures in the composition of delivery services. To develop a fault sentient system that is capable of preventing drastic failures in deliveries, a prediction of failures is required. Due to drone swarms inherent characteristics, like connectivity and internal processing power \cite{yazdinejad2021federated}, several papers studied the use of federated learning in the sky for prediction in different applications. Examples include air quality index predictions \cite{liu2020federated} and drones authentication for privacy and security concerns \cite{yazdinejad2021federated}. We propose to use weighted continual federated learning to accurately predict the soft failing drones and the time of their failures. This prediction would inform the failure-sentient proactive composition of swarm-based drone delivery services.

\section{Swarm-based Drone-as-a-Service Failure Model}

We present a swarm-based drone delivery service failure model. We abstract a swarm carrying packages and travelling in a skyway segment between two nodes as a service (Fig. \ref{motivating}). We formally define an SDaaS service and an SDaaS consumer's request. We then describe the taxonomy of SDaaS failures. The constraints that surround the failure-sentient composition are then described in details. Later, we formulate the problem and the objective function.\\\textbf{Definition 1: Swarm-based Drone-as-a-Service (SDaaS).} An SDaaS is defined as a set of drones, carrying packages and travelling in a skyway segment. It is represented as a tuple of $<SDaaS\_id, S, F>$, where
\begin{itemize}[nosep]
    \item $SDaaS\_id$ is a unique service identifier
    \item $S$ is the swarm travelling in SDaaS. $S$ consists of $D$ which is the set of drones forming $S$, a tuple of $D$ is presented as $<d_1,d_2,..,d_m>$. $S$ also contains the properties including the current battery levels of every $d$ in $D$ $<b_1,b_2, ..,b_m>$, the payloads every $d$ in $D$ is carrying $<p_1,p_2,..,p_m>$, and the current node $n$ the swarm S is at.
    \item F describes the delivery function of a swarm on a skyway segment between two nodes, A and B. F consists of the segment distance $dist$, travel time $tt$,  charging time $ct$, and waiting time $wt$ when recharging pads are not enough to serve $D$ simultaneously in node B.
\end{itemize}
\textbf{Definition 2: SDaaS Request.} A request is a tuple of $< R\_id,\beta, P, T>$, where
\begin{itemize}[nosep]
    \item $R\_id$ is the request unique identifier.
    \item $\beta$ is the request destination node.
    \item $P$ are the weights of the packages requested, where $P$ is $<p_1,p_2,..,p_m>$.
    \item $T$ is the time window of the expected delivery, it is represented as a tuple of the window start and end times $<st,et>$.
\end{itemize}



\subsection{SDaaS Failures Taxonomy}
We propose a novel SDaaS failures taxonomy (Fig. \ref{taxonomy}). 
We classify failures into three levels based on a cause-and-effect relationship. The first level is the drone level. Failures at the drone level would result in failures at the second level, the swarm level. Failures at the swarm level would have an impact on the delivery mission, which is the third level. Below we describe each level in details:

\begin{figure}[htbp!]
\centering
\includegraphics[width=0.9\linewidth]{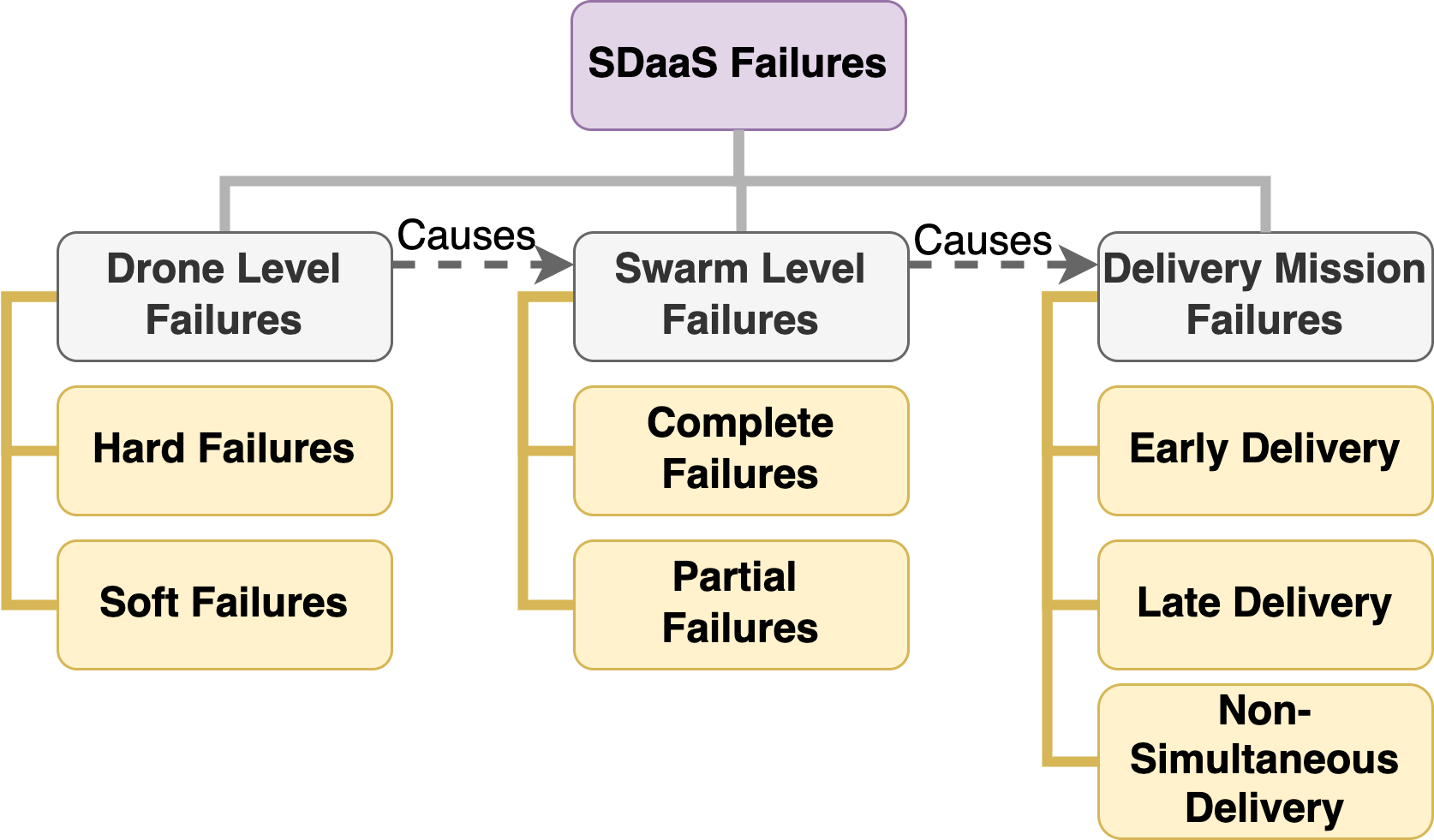}
\caption{SDaaS Failures Taxonomy} 
\label{taxonomy}
\end{figure}

\begin{itemize}
    \item Drone Level Failures: This refers to any failure related to faults in the drones battery, motors, connectivity, sensors, etc. Unlike single drone deliveries, failures at the drone level would impact other non-faulty drones in a swarm since they behave as an atomic entity. Drone level failures are subdivided into two categories based on the impact of the failure on the individual drone:
    \begin{itemize}
        \item Hard Failures: They occur when substantial faults cause the drone to land abruptly or crash catastrophically. Because these types of failures are usually unanticipated and caused by external variables such as wind and bird accidents, they usually necessitate a reactive solution that deals with the delivery after the failure occurs. A hard failure can be defined in terms of the degradation $\delta$ in performance $p$ as follows:
        \begin{equation}
            HF= p-\delta\leq\sigma \qquad \qquad \delta \geq 0
        \end{equation}
        where $\sigma$ is a hard failure predefined threshold.
        \item Soft Failures: They are described as a component, part, subsystem, or system's ability \textit{to continue working even with degraded performance}, i.e., until its reduced performance $p$ falls below a certain predefined threshold $\alpha$. Soft failures are common in components subjected to thermal stress or mechanical degradation \cite{petritoli2018reliability}. A soft failure is defined as:
        \begin{equation}
            SF= \sigma<p-\delta\leq\alpha \qquad \qquad \delta \geq 0
        \end{equation}
        where $\alpha$ is the soft failure threshold.
    \end{itemize}
    \item Swarm Level Failures: This refers to failures that encompasses the swarm which is made of multiple drones. They reflect how many drones within a swarm are failing. The number of failing drones in a swarm would impact the severity of the failures on the delivery mission. The larger the swarm's size, the less reliable it becomes \cite{bjerknes2013fault}.
    \begin{itemize}
        \item Complete Failures: These occur when a swarm can no longer function as a full operating swarm. This could be due to hard failures of a subset or when the number of soft failing drones exceeds a certain threshold $\omega$. A complete failure is defined as:
        \begin{equation}
            CF= (|HF\subset S| \geq 1) \ | \  (|SF \subset S| \geq \omega)
        \end{equation}
        where $|x|$ is the number of elements in $x$. 
        \item Partial Failures: These occur when the number of soft failing drones in a swarm is below the threshold $\omega$. In a partial failure, no hard failing drones exist as the swarm is still capable of functioning as a full swarm. A partial failure is defined as follows: 
        \begin{equation}
            PF= (|HF\subset S| = 0) \ \& \  (|SF \subset S| < \omega)
        \end{equation}
    \end{itemize}
    \item Delivery Mission Failures: In SDaaS, a delivery mission is successful $SD$ if the packages are delivered \textit{simultaneously} within the consumer's \textit{specified time window} \cite{alkouz2021reinforcement}.
    \begin{equation}
    \small
            SD= (AT(d_1 -d_n) \leq \rho) \& (st<AT(d_1 -d_n)<et)
    \end{equation}
    
    where $AT$ is the arrival times of the drones to a destination and $\rho$ is the time window threshold between arriving drones. A delivery mission failure occurs when any of the aforementioned conditions is broken. Delivery mission failures are subdivided to:
    \begin{itemize}
        \item Early Delivery: If any or all packages arrive before the consumer's specified time window, the delivery mission is considered a failure. This is owing to the fact that a consumer may be away when the products arrive, putting the packages at danger of theft or damage due to weather.
        \item Late Delivery: Consumers are generally dissatisfied with late deliveries. This is due to the fact that some goods may be urgent. As a result, we consider late deliveries to be failed deliveries.
        \item Non-Simultaneous Delivery: The promise of Swarm-based drone delivery services is to deliver multiple packages together. This is important specially for packages that are considered one item, eg. PC or furniture parts, synergic drugs, etc. Moreover, drone deliveries usually occur on a building's rooftop, and packages need to arrive together for the consumer's convenience \cite{wang2009web}. Therefore, if packages arrive separately, then it is considered a non-simultaneous delivery failure.
    \end{itemize}
\end{itemize}
In this paper, we focus on minimizing the impact of \textit{soft failing drones} on the swarm and delivery mission.
\begin{figure*}[htbp!]
\centering
\includegraphics[width=0.9\textwidth]{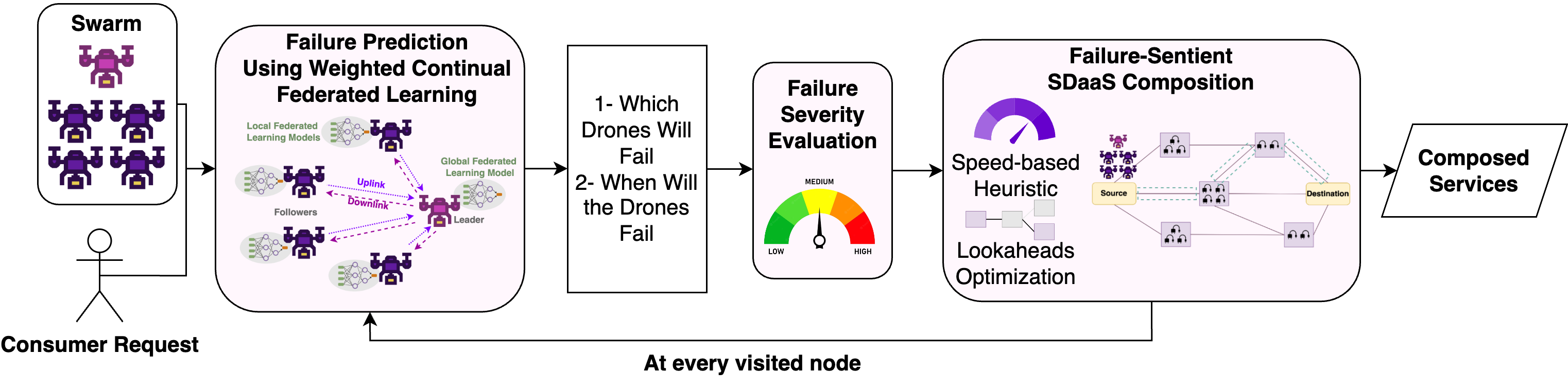}
\caption{Failure-Sentient SDaaS Framework} 
\label{framework}
\end{figure*}

\subsection{SDaaS Failure-Sentient Composition Constraints}
There are several constraints surrounding the optimal composition of swarm-based drone delivery services under failure. Some are related to the inherent specification of the drones and others related to the delivery environment. Below we state these constraints:
\begin{itemize}
    
    \item \textit{Varied Severity and Times of Degradation:} Drones in a swarm typically have various operation start dates and flight histories. A drone, for example, might have been more frequently used to transport packages in windy situations or to carry heavier payloads than other drones. As a result, each drone's degradation time and rate is unique. Based on the drone's past data, our solution must precisely predict the degradation time and severity.
    \item \textit{Different Energy Consumption Rates:} Drones in a swarm carry varying payloads, leading to varying energy usage rates. Furthermore, drone swarms may fly in formations \cite{liu2022constraint}, with each drone's energy consumption influenced by its position within the formation \cite{alkouz2020formation}\cite{guo2023drone}. Furthermore, if multiple drones degrade at various rates, the energy consumption rate would be impacted. As a result, our solution must compose the path to account for these variances and prevent soft failures from becoming hard failures.\looseness=-1
    \item \textit{Speed Trade-off:} The speed of the drone has a significant impact on its energy consumption. Increasing the speed puts additional strain on the motors, increasing the rate of energy consumption. Furthermore, when speed increases, so does air resistance, putting more strain on the battery \cite{dukkanci2021minimizing}. Similarly, decreasing a drone's speed minimises its energy consumption. As a result, it is difficult to know when to increase the speed to shorten travel time and when to lower the speed to reduce energy consumption and preserve charging time. Previous research on the composition of drone delivery services assumed that the drones would travel at a constant speed \cite{alkouz2020formation}\cite{alkouz2022flight}. 
    \item \textit{SDaaS Environment Constraints:} Due to the large number of drones in a swarm, there may be instances when the number of drones exceeds the number of recharging pads at a node. Therefore, the waiting cost due to sequential charging will need to be considered. 
     \item \textit{Failures Not Occurring as Predicted:} Predicting a soft failure is not always accurate. This is due to the fact that a degradation in performance is dependent on numerous variables. As a result, our solution should handle situations where failures do not occur as predicted and how this \textit{new information} would be integrated to predict the next soft failures. 

\end{itemize}

\subsection{Problem Formulation}
Given an atomic swarm $S$ made of $n$ drones $D={d_1, d_2,...d_n}$ and a consumer request $R_i$ with expected delivery time window $<st,et>$, the problem is formulated as predicting soft failing drone/s and composing the optimal set of services to deliver the packages together under failure within the request specified time window considering all the aforementioned constraints. The degradation in performance should have a minimal influence on delivery such that the consumer is not aware of it. The composition problem lies under the family of combinatorial optimization problems, where we try to find an optimal solution from a finite set of solutions. This is an NP-Hard problem since there is no quick solution and the complexity of calculating the route increases as we add more nodes and drones to the problem due to the increased number of permutations and combinations.


\section{Failure-Sentient SDaaS Framework}

The failure-sentient composition framework consists of three main modules. The first module is the prediction module using weighted continual federated learning. This module predicts which drones will fail, time of their failures, and their uptime after failures. The second module computes a failure severity score on the drone and swarm levels. The third module uses the prediction output and severity evaluation to compose the best failure-sentient services that will meet the consumer's needs. As shown in Fig. \ref{framework} the prediction, evaluation, and composition phases occur at each node where the swarm lands to recharge until it reaches its destination. The process is iterative at each visited node since the framework is proactive and \textit{failures may not always occur as predicted}, necessitating a new prediction based on recent history and composition based on the new prediction.

\begin{figure*}[htbp!]
\centering
\includegraphics[width=0.9\textwidth]{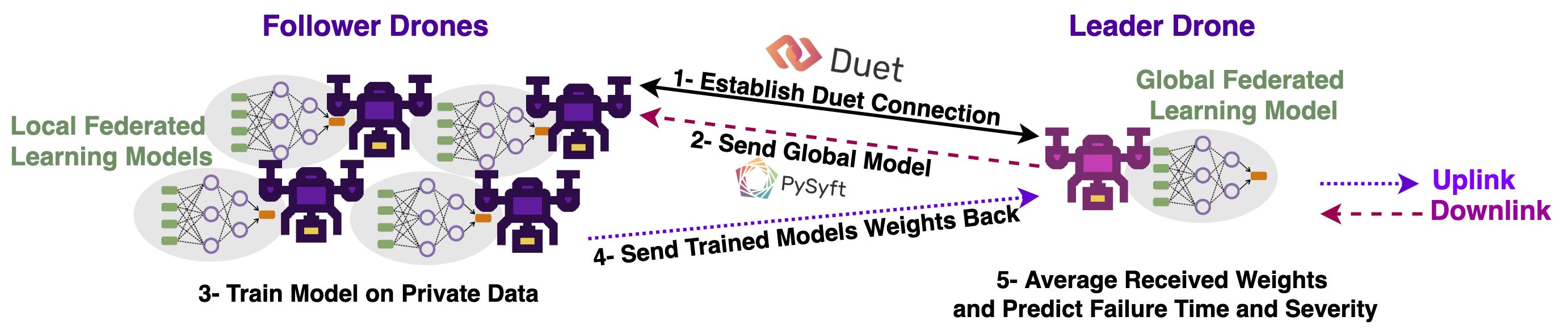}
\caption{FL Workflow for Failure Prediction} 
\label{FederatedLearning}
\end{figure*}

\subsection{Failure Prediction Using Weighted Continual Federated Learning}
Given a swarm of drones, the goal of this module is to predict the drones whose performance would degrade and estimated time of the degradation. As described earlier, the drones in a swarm are homogeneous in type and capabilities but heterogeneous in delivery history. A drones delivery history consists of the carried payloads, the weather conditions it travelled in, the distances it travelled, the drones lifetime, etc. Therefore, according to the history, each drone would typically face a different degradation time and rate of performance. In this paper, we focus mainly on the degradation in power consumption. A degradation in power consumption occurs when a drone consumes energy ($ec$) more than what is expected ($ex$), i.e. $ec>=ex+\epsilon$. Where $\epsilon$ is the threshold where beyond it, the drone is considered to be soft failing. For this purpose, we propose to predict the failures using weighted continual federated learning.

Federated learning (FL) is a machine learning technique that trains an algorithm across multiple decentralized devices holding local data samples, without exchanging them \cite{liu2020federated}. Since drones in a swarm generate new data as they fly that may be useful in prediction, having a centralised learning by sending the data to the ground base station will have high latency that may slow down the learning process. Moreover, in the future, drones from different owners may be utilized to create a swarm service. In this case, the privacy of each drones data is crucial and a decentralized learning would be necessary. Since the drones generate data as they fly, we propose to divide a drones historical data into old and new history. The old history is the data generated from previous flights since the drone was initially deployed. The new history is the data generated in the current flight till the visited node. Therefore, the data from the current flight may provide more valuable insights to the current drone status helping in more accurate failure predictions. Therefore, the learning would be: 1) \textit{weighted}, where the new history would have higher weight in prediction than the old history, and 2) \textit{continual}, where the new history is incremented at every visited node. Since machine learning is a time and energy extensive task, we propose that the learning occurs when the drones are idle and spending time charging at the intermediate nodes.


Consider a swarm of drones flying as shown in Fig. \ref{FederatedLearning}. The swarm is made up of a leader $l$ and a group of $f$ followers. The drone swarm captures data representing the new history while flying. FL is performed at the source or when the swarm lands at intermediate nodes to predict soft failures in the following segment/s. 
For the implementation, we consider FL with PySyft and PyTorch \cite{ziller2021pysyft}. PySyft is a Python library built for secure and private deep learning. PySyft decouples private data from model training using FL within the main machine learning frameworks like PyTorch. 

Fig. \ref{FederatedLearning} illustrates the steps involved in the FL failure prediction. In step 1, while the swarm is at the source, a Duet connection gets established between followers and leader drone.  Duet is a peer-to-peer tool within PySyft designed so that learning can be made on data that is inaccessible or tightly controlled. After the connection is established, a global model is distributed to the followers drones (step 2). We use a linear regression model to predict the time to failure of each drone and for how long could each drone travel after failure. 
In step 3, the follower drones train the local models on their own private historical data. In step 4, the follower drones send the models weights back to the lead drone. In step 5, the lead drone averages the weights received from all drones. The lead drone predicts the time to failures and uptimes after failures that are used in the next modules to compute failure severities and compose the best service (step 5). After that at every visited node, steps 1-5 are repeated except at steps 3 and 4 the follower drones use the model to predict on two sets of data, i.e. the old history and the new history. 
Before training and returning the weights to the lead drone, more weight is put on the new history. The new history is given more weight by considering each data point multiple times based on the weight specified. The weights are determined empirically and evaluated in the experiments section. The output of this module is the time to failure and uptime predicted for each drone. These will be used as an input to the next module, i.e. the severity evaluation.\looseness=-1

\subsection{Failure Severity Evaluation}
\begin{figure}[htbp!]
\centering
\includegraphics[width=0.9\linewidth]{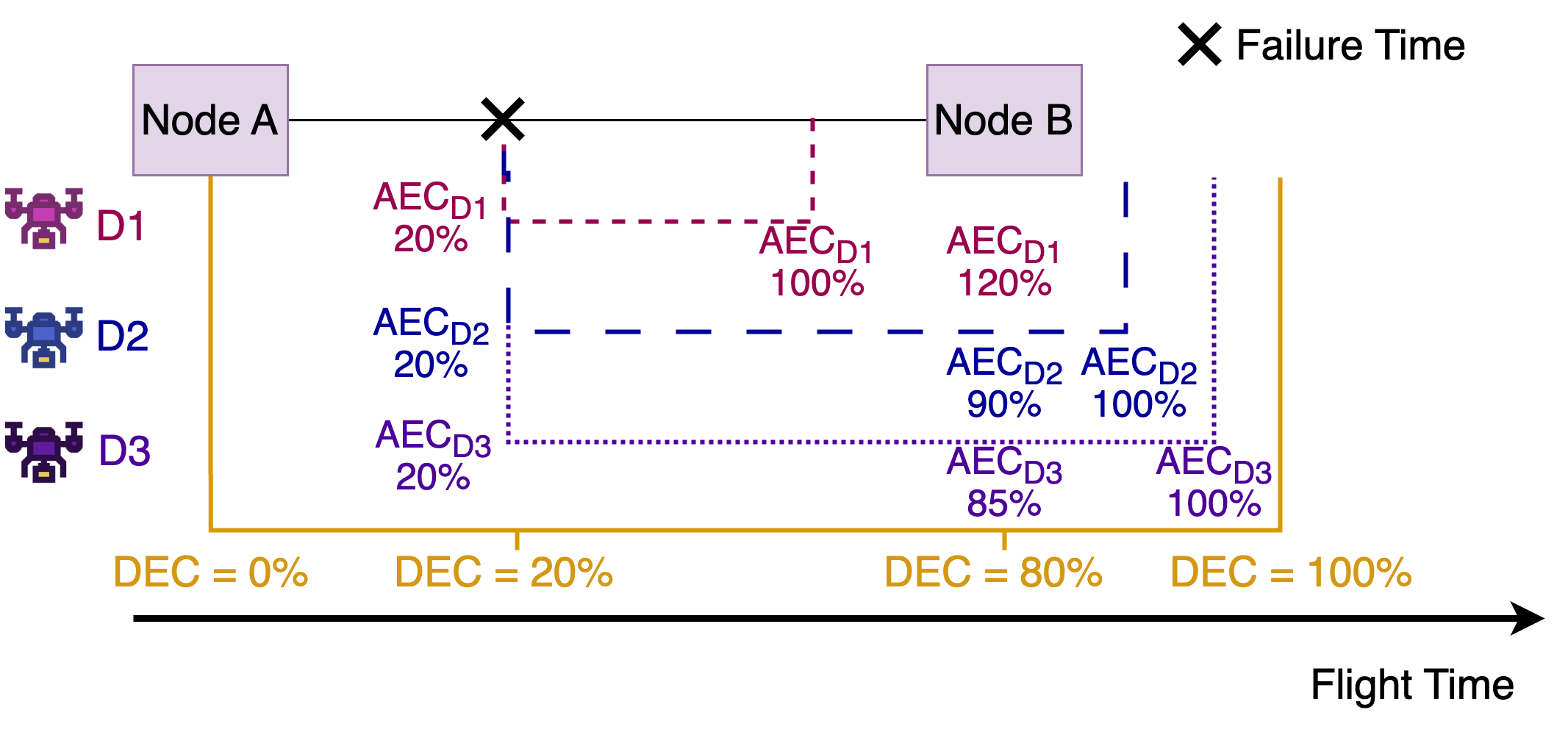}
\caption{Scenarios with Varying Durations of Failure Uptime} 
\label{severity}
\end{figure}

\begin{figure*}[htbp!]
\centering
\includegraphics[width=0.75\textwidth]{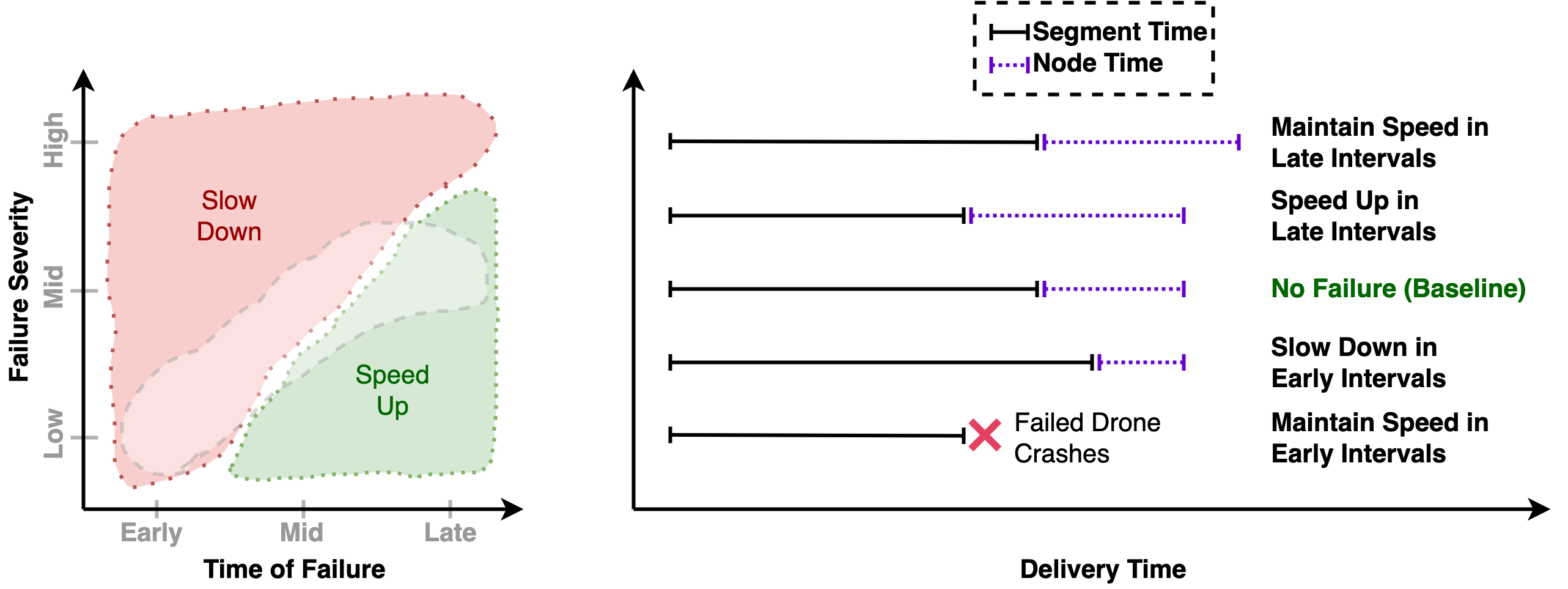}
\caption{Speed Heuristic for Failure-Sentient Composition} 
\label{speeddecision}
\end{figure*}
In this module, the severity of the failure on the drone level and the swarm level is computed. This module takes as an input the predicted Failure Time ($FT$) and the uptime, i.e. the time the drone can still operate after a soft failure, for each drone. To compute the severity score at the drone level, i.e. for each drone, we use the Default Energy Consumption ($DEC$) when no failure exists and the Actual Energy Consumption ($AEC$) rates after failure. Fig \ref{severity} illustrates examples of homogeneous soft failing drones at time $t1$ and their uptime after failure. The figure also shows the energy consumed in percentage of the failing drones versus the default energy consumption if no failure occurred. We assume that the calculation of $DEC$ is based on the theoretical consumption model proposed by Stolaroff et al. \cite{stolaroff2018energy}. The model expresses the overall energy per meter (J/m) as follows:

\begin{equation}
Epm= \frac{T(v_{a}sin \alpha) +v_{i}}{v_{a}\eta}
\label{epm}
\end{equation}

where $T$ is the thrust of the drone, taking into account all its physical components: (i) drone body, including the airframe, propellers, motors, sensors, GPS, and, if applicable, a camera, (ii) drone battery, and (iii) payload. The airspeed of the drone relative to air is denoted as $v_{a}$ (m/s), while $\alpha$ represents the angle of attack, which is the angle in radians of the airspeed to the drone rotor. The induced speed $v_{i}$ (m/s) is the change in speed of the air after it flows through the drone, and the battery power transfer efficiency $\eta$ is provided by the manufacturer.

The $DEC$ in percentage is obtained by converting the Epm value, as illustrated in the figure. For instance, if we consider the DJI Phantom 3 drone model and assume that the Epm equation yields a consumption of 20J/m, then the maximum distance that the drone can cover based on the battery specifications is given by the total energy capacity divided by the energy consumption rate per unit distance. Therefore, the maximum distance the drone can cover is 245145.6 (J) / 20 (J/m) = 12257.28 (m). The value of 245145.6 J was obtained by utilizing the specifications given by DJI for the Phantom 3 battery. These specifications state that the battery has a capacity of 4480 mAh and operates at 15.2 V, resulting in a total energy capacity of:
E = 4480 mAh x 15.2 V = 68.096 Wh = 68.096 x 3600 J = 245145.6 J.

The percentage values for $AEC$ are calculated based on the predicted time to failure and uptime after failure from the preceding module. We use $DEC$ and $AEC$ values to compute the Drone Level Severity ($DLS$) using the equation below:
\begin{equation}
    DLS= 1- \frac{DEC_{node}-DEC_{FT}}{AEC_{node}-AEC_{FT}}
\end{equation}

where $DEC_{node}$ and $AEC_{node}$ are the actual and default energy consumption percentages for the drone at the next node (Node B) and $DEC_{FT}$ and $AEC_{FT}$ are energy consumption percentages at the failure time. In the example, the $DLS$ value for D1 is 0.4. A threshold value ($\sigma$) is needed to determine if this failure is hard or soft. In our context, a drone is considered to have a hard failure if it is not able to reach the next node due to excessive energy consumption. We define $\sigma$ as:
\begin{equation}
    \sigma = 1- \frac{DEC_{node}}{100}
\end{equation}
assuming 100 is the maximum battery percentage a drone could have. The threshold value in the above example will be 0.2. If $DLS$ is higher than the $\sigma$ we consider the drone to have a hard failure. Moreover, if two $DLS$ values are less than $\sigma$, both are considered soft failures and the drone with a higher $DLS$ value is facing a more severe failure. In the example, D2 ($DLS=0.14$) is facing a more severe failure than D3 ($DLS=0.08$).

As described earlier, a drone or a set of drones could have an impact on the full operation of the whole swarm. Therefore, we need to compute a swarm level severity of the failure from the individual drones. We use the equation below to compute the Swarm Level Severity ($SLS$):

\begin{equation}
    SLS = \frac{\sum\limits_{i=1}^{|SF| + |HF|} DLS }{|SF| + |HF|}
\end{equation}

where the number of soft and hard failing drones in the swarm is represented by the sum of $|SF|$ and $|HF|$. If we consider a swarm of five drones, having two drones with a high $DLS$ is riskier than having four drones with low $DLS$ values. The $SLS$ values range from 0 to 1 and are divided into thirds to categorize the severity level as low, mid, or high. By computing the $SLS$ value, we can determine appropriate strategies for mitigating hard failures and extending battery life, as detailed in the following subsection. 

\subsection{Failure-Sentient SDaaS Services Composition}

In this module, we propose a speed-based heuristic approach to compose the optimal set of services and navigate the swarm between the source and destination nodes. While the swarm is at the source, the prediction module provides the expected time of soft failure and the severity of power degradation. If no failure exists, the swarm would select the neighbour that results in the least travel time $tt$ and node time $nt$. The node time consists of the charging times $ct$ of the drones and the waiting times $wt$ caused by sequential charges as explained earlier. Once a node is selected, the swarm traverses to that node. At the node, while the swarm is charging, prediction using the old history and the new history from the traveled segment takes place. We assume a swarm traverses at a fixed speed recommended by the drone manufacturer, e.g 105km/h. If a single drone failure is predicted at a segment, a speed-based heuristic takes place to control the speed of the swarm. The speed-based heuristic is dependant on two factors as shown on the axes in Fig. \ref{speeddecision}, namely the time of failure and the severity of the degradation. We propose to divide each skyway segment into three equal intervals by time (assuming the fixed base speed). We divide a segment into intervals because the predicted time of failure would be inexact but would most likely fall within one of these intervals. Based on the severity and time of failure, we explain when a swarm would slow down or speed up and the intuition behind each choice. The right side of Fig. \ref{speeddecision} describes the impact of different speed decisions against the baseline scenario where no failure is assumed to occur. The idea is that even with soft failures, the overall delivery time is not impacted enough that the consumer notices. The speed decisions based on different degradation scenarios are described below:

\begin{itemize}
    \item Slow Down: If a failure occurs at the first (early) interval this means that there is still a long distance to travel on the segment. Therefore, maintaining the current base speed may cause the battery to be depleted before arriving to the next node. Therefore, slowing down the the drones would give higher chances for the swarm to survive by consuming less energy avoiding a hard failure due to power depletion. The drones would slow down to a speed where their energy consumption rate would take them to the next node safely based on the severity of the degradation. Slowing down healthy drones would also mean that they consume less energy which would result in less node times due to charging and waiting. Similarly, if a failure occurs at the mid or last interval and the degradation in power is high, then the drones would slow down. This is because even if the distance to travel is small the risk of hard failures are high due to severity. 
    \item Speed Up: Speeding up would typically result in shorter travel times but more energy consumption that may lead to added charging times. Therefore, if a degradation severity is low or medium and the failure occurs at the last interval of a segment, the swarm would speed up as there is low risk of catastrophic failure. This is because the distance left to travel would be small and there would be low chances of full battery depletion. A swarm would speed up to what takes it to the next node according to the energy consumption model of the used drone given the payload, speed, and degradation. 
    \item Maintain Speed: In some situations, represented by the grey area in Fig. \ref{speeddecision}, the speed decision is unclear. In these cases, if it overlaps with the speed up area, the swarm would try to speed up if the battery permits. Otherwise, it would maintain its speed. Similarly, with areas overlapping with speed down, the swarm tries to maintain speed and would only slow down if one or more drones are not able to reach the next node when computing the expected battery level. 
\end{itemize}
If \textit{multiple failures occur} at a segment, the decision of speeding up or slowing down (y-axis in Fig. \ref{speeddecision}) depends on the $SLS$ value computed in the previous subsection. \textit{The time of speeding up or slowing down occurs at the time of failure of the most severely degraded drone} (x-axis in Fig. \ref{speeddecision}). Once a swarm is at the next node, the drones would recharge their batteries fully. The total delivery time is incremented by the travel time of the segment and the time spent at the node recharging. While recharging, the prediction is done again but this time by using the new data generated from the travel at the segment in addition to the old data with more weight on the new history. Predictions on the new history would reflect the real status of the drone as well as serve as an input if failure does not occur as expected at previous segments. The steps are repeated until all drones are at the destination node. If a swarm at the second last node is expected to arrive at the destination earlier than the consumer specified time, it would wait before traversing the last segment. The final delivery time is the sum of all segments travel times and node times \cite{lee2021package}.

\subsubsection{Lookaheads optimization}
\begin{figure}[htbp!]
\centering
\includegraphics[width=0.9\linewidth]{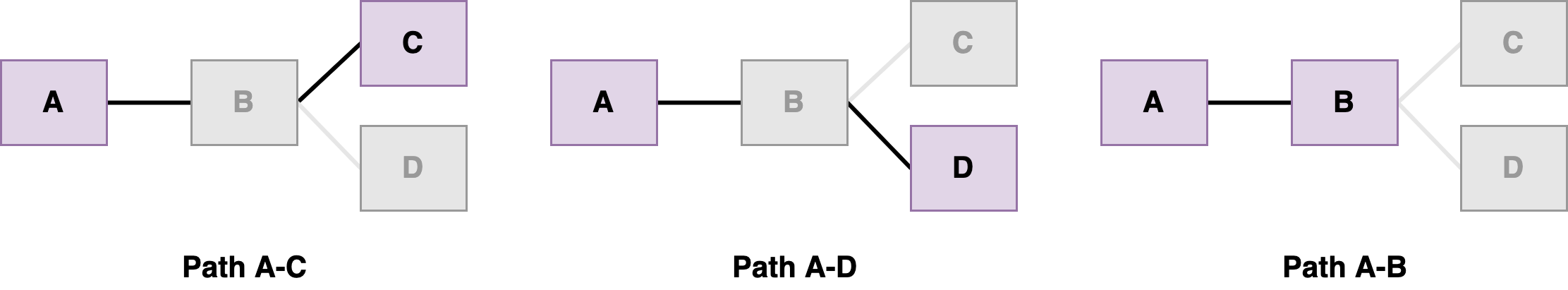}
\caption{Path Composition Scenarios with Lookaheads} 
\label{lookaheads}
\end{figure}

The proposed approach composes one segment at a time. However, in some instances, a swarm is able to travel multiple segments before requiring any recharge stops. Therefore, we propose to use a lookaheads optimization approach to reduce the number of recharging stops. In this case, when lookahead = 1 for example, the directly connected nodes and the 2\textsuperscript{nd} level connected nodes are considered in the composition. As shown in Fig. \ref{lookaheads}, depending on the reachability and battery capacity, the composed segment would either be A-C or A-D without stopping at B, or A-B if C and D are unreachable due to power constraints.

\section{Experiments}
\label{experiments}

In this section, we evaluate the performance of the proposed modules in the framework. We evaluate the performance of the prediction module in terms of accuracy. We then assess the severity evaluation and composition modules in terms of successful requests, delivery, and execution times. We run all the experiments using Python on a MacBook Pro, Apple M1 chip, 16 GB memory, and 8 cores. Two real datasets are used for the evaluation of the framework. The first dataset is the ALFA dataset that includes data on drones faults \cite{keipour2021alfa}. The second dataset is an urban road network dataset to construct and mimic a skyway network \cite{karduni2016protocol}. Below we describe each dataset and the data engineering process performed on each:
\begin{itemize}
    \item ALFA dataset: 
    The ALFA dataset provides fixed-wing drones flight logs in real test flights with actual control component failure. Experimental real-world flights were performed at the airport in Pittsburgh, the United States. The ALFA dataset provides five types of flight logs: safe flight, a flight under failures at the aileron, rudder, engine, and the elevator. The dataset contains data of the time of fault and for how long did the drone fly afterwards with the fault (uptime). The flight logs recorded in UAVs include a wide range of features. We use the following features to aid in the prediction of severity and times to failures: velocity (x,y,z), angular velocity (x,y,z), linear acceleration (x,y,z), magnetic field (x,y,z), fluid pressure, temperature, altitude error, airspeed error, tracking error, and distance between ideal location and current location. After selecting the key features, we used feature engineering techniques to convert the flight logs into feature vectors. The two processes are feature scaling and timestamp range merge. Feature scaling is needed as each feature has different range of values. We perform range scaling or normalization to unify the range of values for each feature. We use standardization to normalize the dataset according to the following equation:
    \begin{equation}
        z= \frac{x-mean}{std}
    \end{equation}
    
    where $std$ refers to the standard deviation of a feature data. The mean and standard deviation values are with respect to the dataset as a whole. After normalization, we range merge the datapoints into a timestamp of 1 second. The reason is that each feature has a different reported timestamp. This could be due to the fact that they are collected using different sensors. Therefore, we take the average value of datapoints for each feature at a one second interval. This data engineering process results in a unified timestamp and feature scale for all features. This processed feature vector serves as the input to the prediction module to predict the time to failure and the uptime after failure. We assess the severity of the failure according to how much time the drone was able to fly after the failure occurrence (uptime).
    
    \item The second dataset used in the experiments is a New York city urban road network dataset. The data consists of the city graph edge list \cite{karduni2016protocol}. To mimic a skyway network, we used a sub-network of 208 connected nodes for the experiments. Each node was randomly assigned a different number of charging pads. We then generate a set of 1700 requests with different source and destination nodes. For each request, we synthesize payloads with a maximum of 5 packages and a maximum weight of 2.5kg. Based on the distance traveled, and the payload carried, all power consumption calculations are made for the DJI Phantom 3 drone model.
\end{itemize}

\begin{figure}
  \centering
  \includegraphics[width=0.9\linewidth]{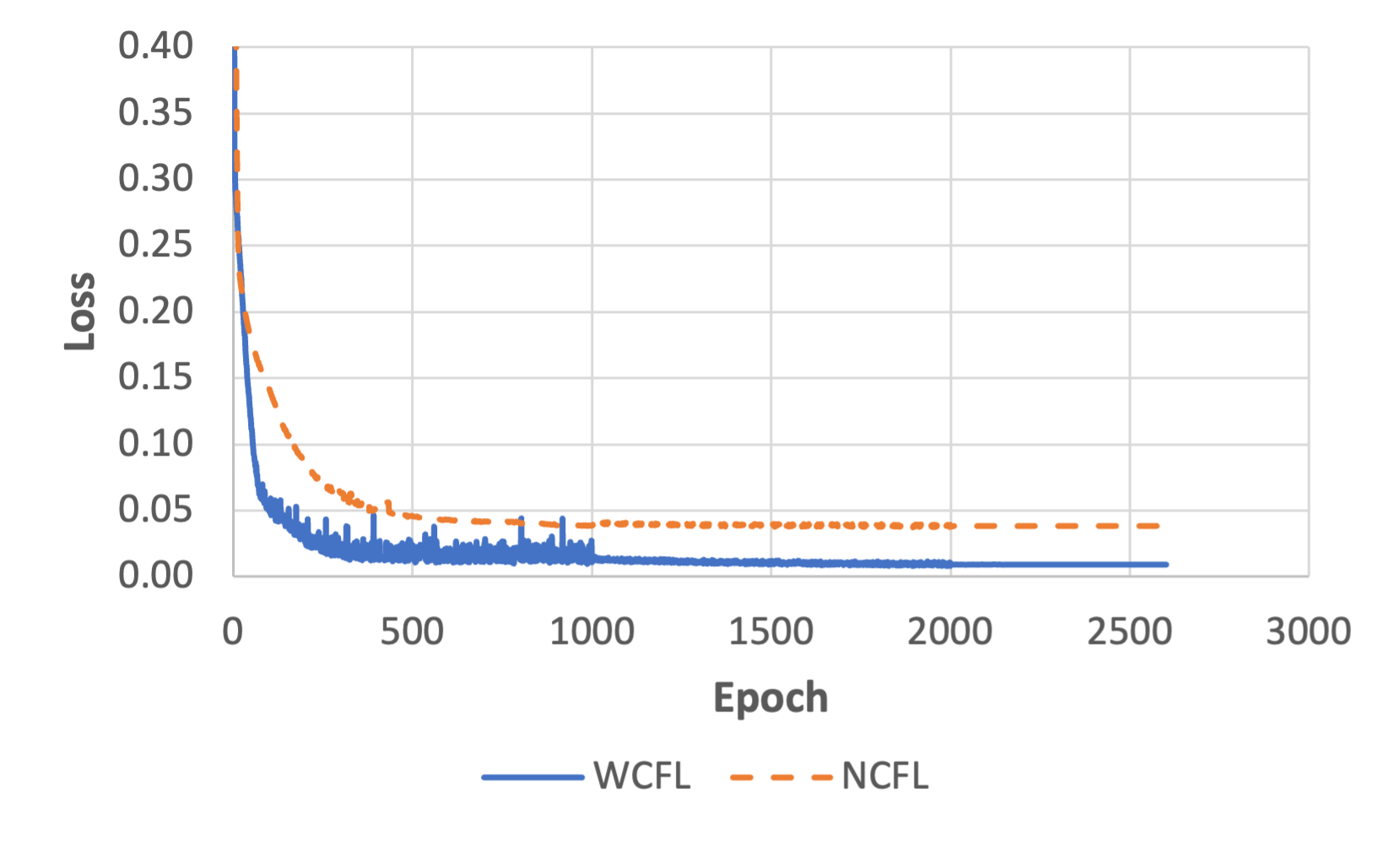}
  \captionof{figure}{Loss using WCFL Prediction}
  \label{fig:loss}
\end{figure}

In the first experiment, we evaluate the accuracy of the prediction module. Using the ALFA dataset, we compare the weighted continual approach to a non-continual learning approach. For the weighted continual federated learning (WCFL) approach we put more weight on the newly generated data as described earlier. We do this by considering each data point from the new data multiple times depending on the set weight. Fig. \ref{fig:loss} shows how the WCFL with weight 2, where each data point of the new history is considered twice, outperforms the non-continual federated learning approach (NCFL) in prediction. The loss in the y-axis is calculated using the mean absolute error between each element in the predicted $x$ and the observed $y$ in the test set. This observation proves that emphasising the new data when learning helps in more accurate prediction since it reflects what could happen in the near future more. However, as shown in the figure, the WCFL takes more time to converge since each new data point is considered more than once. 

\begin{figure}
  \centering
  \includegraphics[width=0.9\linewidth]{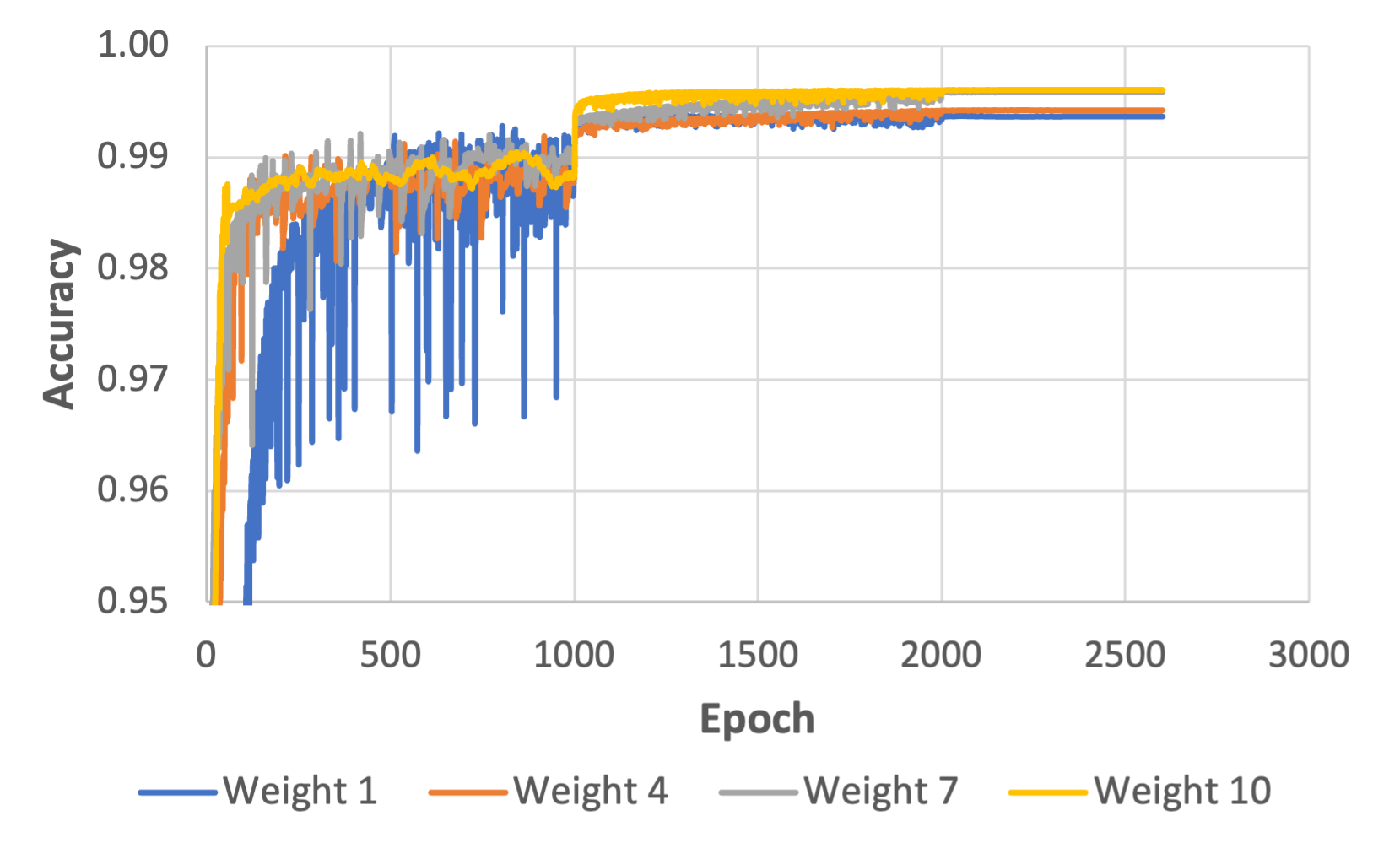}
  \captionof{figure}{Accuracy with Different Weights in WCFL Prediction}
  \label{fig:weights}
\end{figure}

In the second experiment, we evaluate the effect of different weights of the WCFL. A weight of 1 means that each data point in the new history is considered once as is with the old history. A weight of 2 means that each data point is considered twice compared to the data points of the old history which are still considered once. Fig. \ref{fig:weights} illustrates the effect of increasing the weights on the accuracy of the failure predictions. As shown in the figure, the accuracy increases slightly as the weight increases. However, it reaches to a point where the increase of accuracy is very minor. Moreover, with increased weights the accuracy tends to stabilise earlier and that could be because the dataset size increases with higher weights as each data point is considered many times. However, increasing the weight and considering more data points means a slower execution time. Therefore, a trade-off between required accuracy and computation time needs to be considered. At epoch 1000, a notable improvement in the training accuracy is observed due to the systematic reduction of the learning rate by orders of 10 (e.g., 1e-2, 1e-3, 1e-4) at every 1000th epoch. The rationale for adopting a large initial learning rate is to obtain an average model expeditiously, followed by a gradual reduction of the learning rate to refine the model's accuracy

In the rest of the experiments, we evaluate the failure-sentient composition module using the failure times and severities predicted from the first module. We evaluate our proposed speed-based heuristic approach and the lookaheads optimization with two other approaches. First, we evaluate against an exhaustive search approach where the speed-based heuristic is run on top 100 shortest paths by distance between the source and the destination. To avoid computational infeasibility with larger networks, we implemented the top 100 cap on the exhaustive search to prevent it from searching all feasible paths. Second, we evaluate against the safer yet greedy approach of slowing down the drones at all times to conserve energy and protect the drones from hard failures.

\begin{figure}
  \centering
  \includegraphics[width=0.9\linewidth]{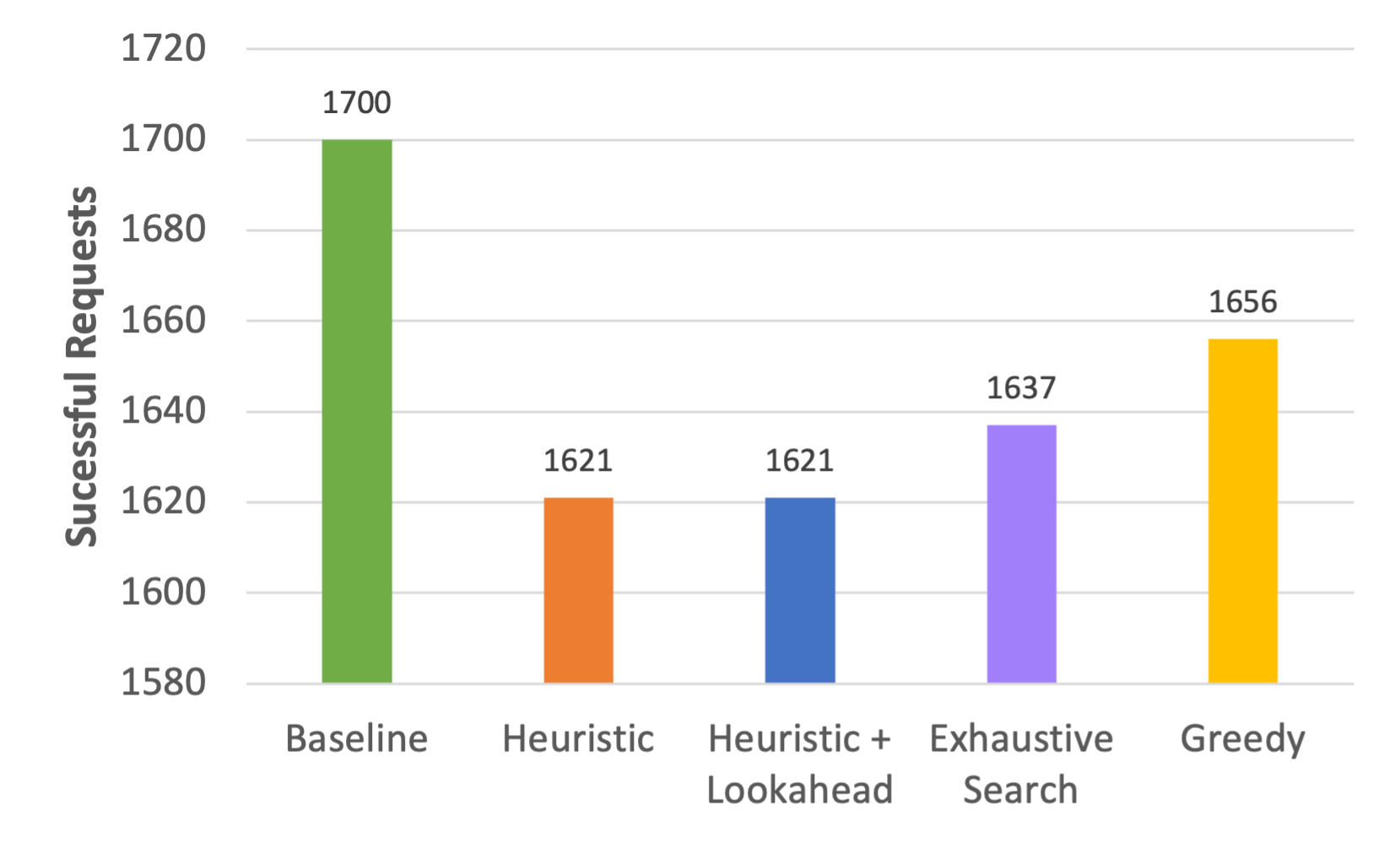}
  \captionof{figure}{Number of Successful Requests}
  \label{fig:successful}
\end{figure}

\begin{figure}
  \centering
  \includegraphics[width=0.9\linewidth]{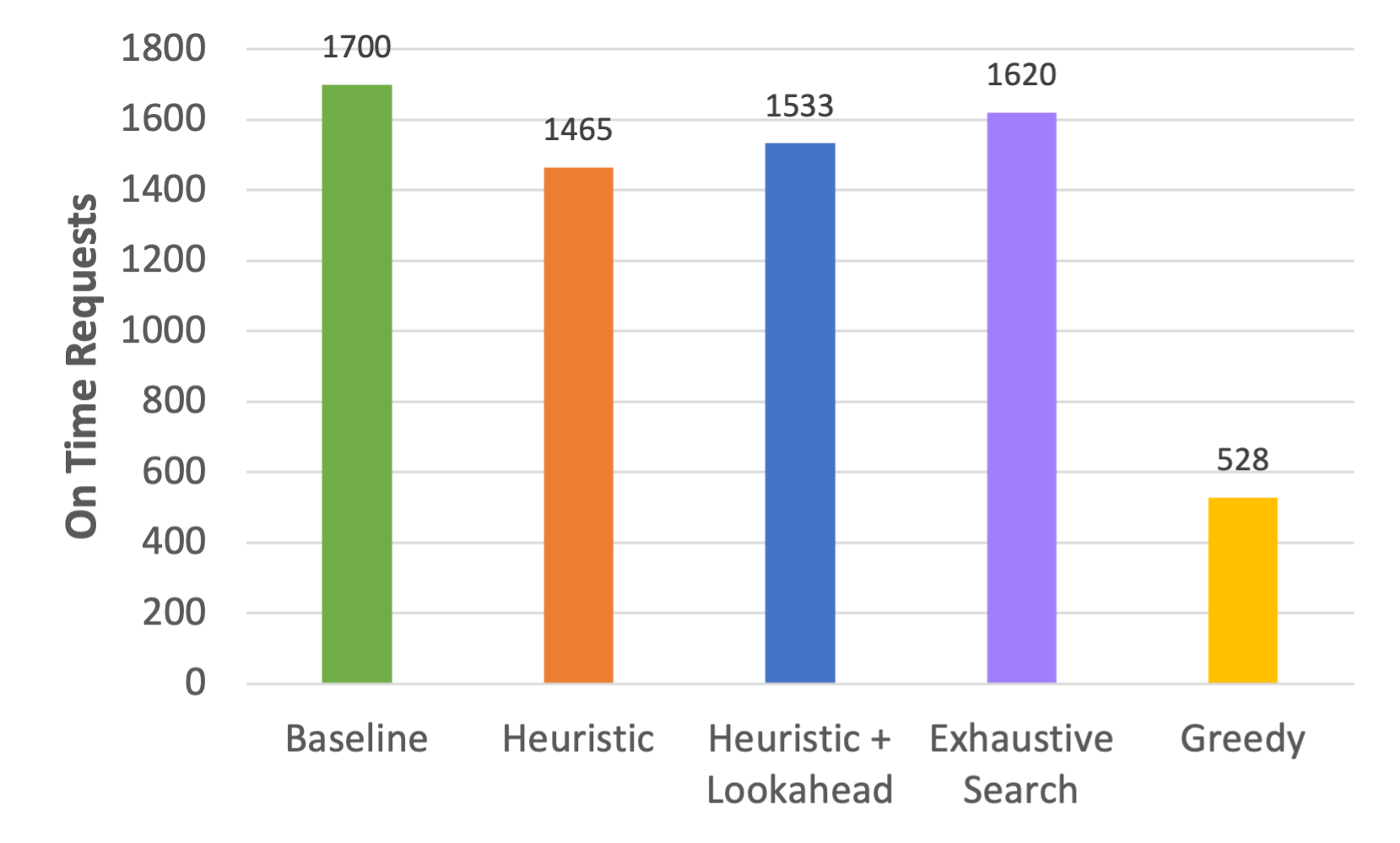}
  \captionof{figure}{Number of On Time Requests}
  \label{fig:ontime}
\end{figure}

For the third experiment, as shown in Fig. \ref{fig:successful}, we measure the number of successful requests of all different approaches. In this case, a successful request is one that, even in the event of a failure, could still deliver the packages to the destination before the batteries depletion. As shown in the figure, the baseline represents a scenario where no failures occur. The greedy approach is able to deliver more orders under failure. This is because it is following a very safe yet slow approach. The exhaustive search approach is the second best in terms of number of delivered packages as it looks into all possible paths with higher probability in finding a successful one using the speed-based heuristic. Our proposed heuristic and the lookaheads optimization were delivering less number of packages as there is still some risk associated with speeding up and consuming more battery when the length of the segment is very long. Although the greedy algorithm delivers more requests safely to the destination, due its safe approach of slowing down it delivers the least requests on time as shown in Fig. \ref{fig:ontime}.

\begin{figure}
  \centering
  \includegraphics[width=0.9\linewidth]{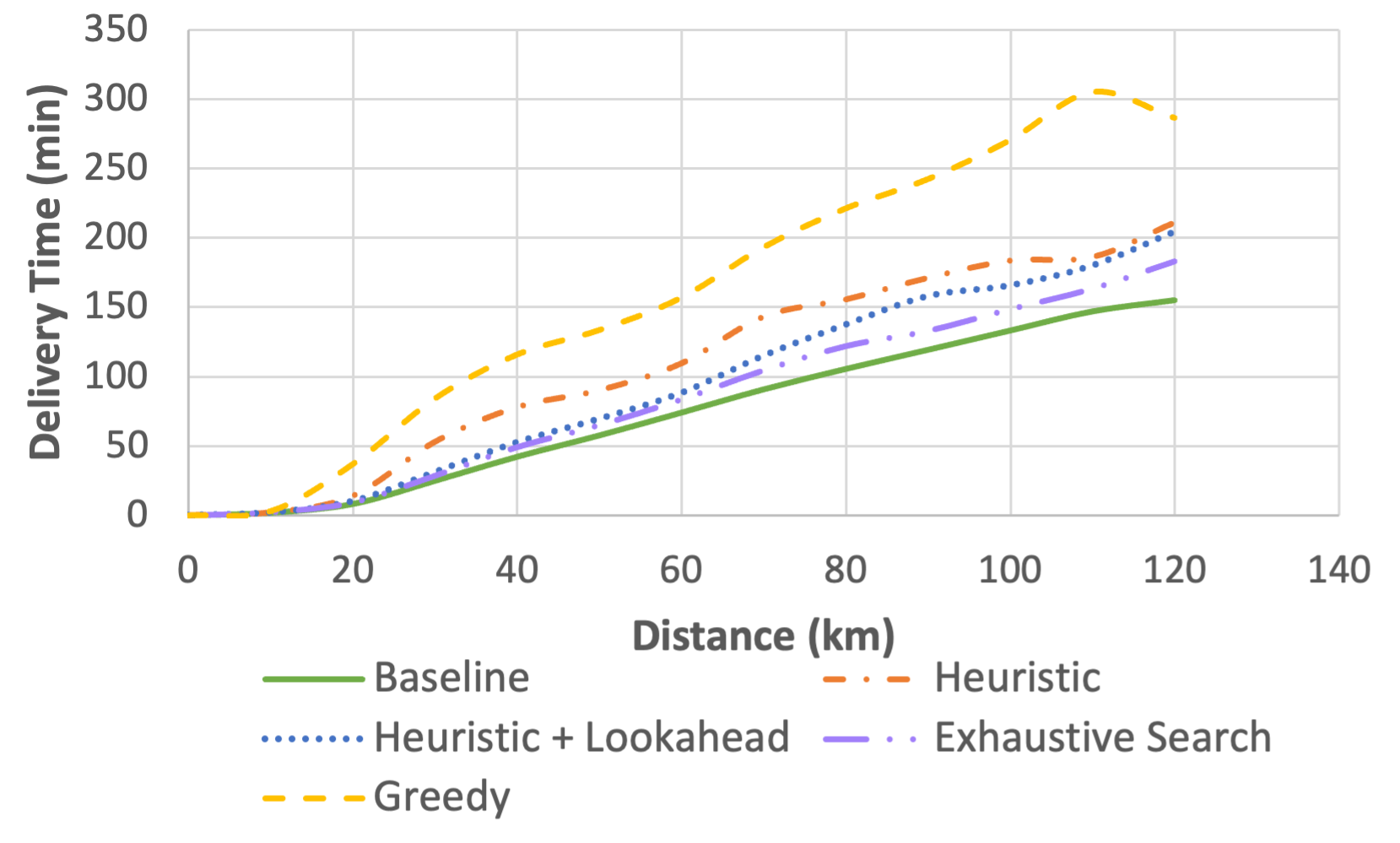}
  \captionof{figure}{Average Delivery Times}
  \label{fig:deliverytime}
\end{figure}

\begin{figure}
  \centering
  \includegraphics[width=0.9\linewidth]{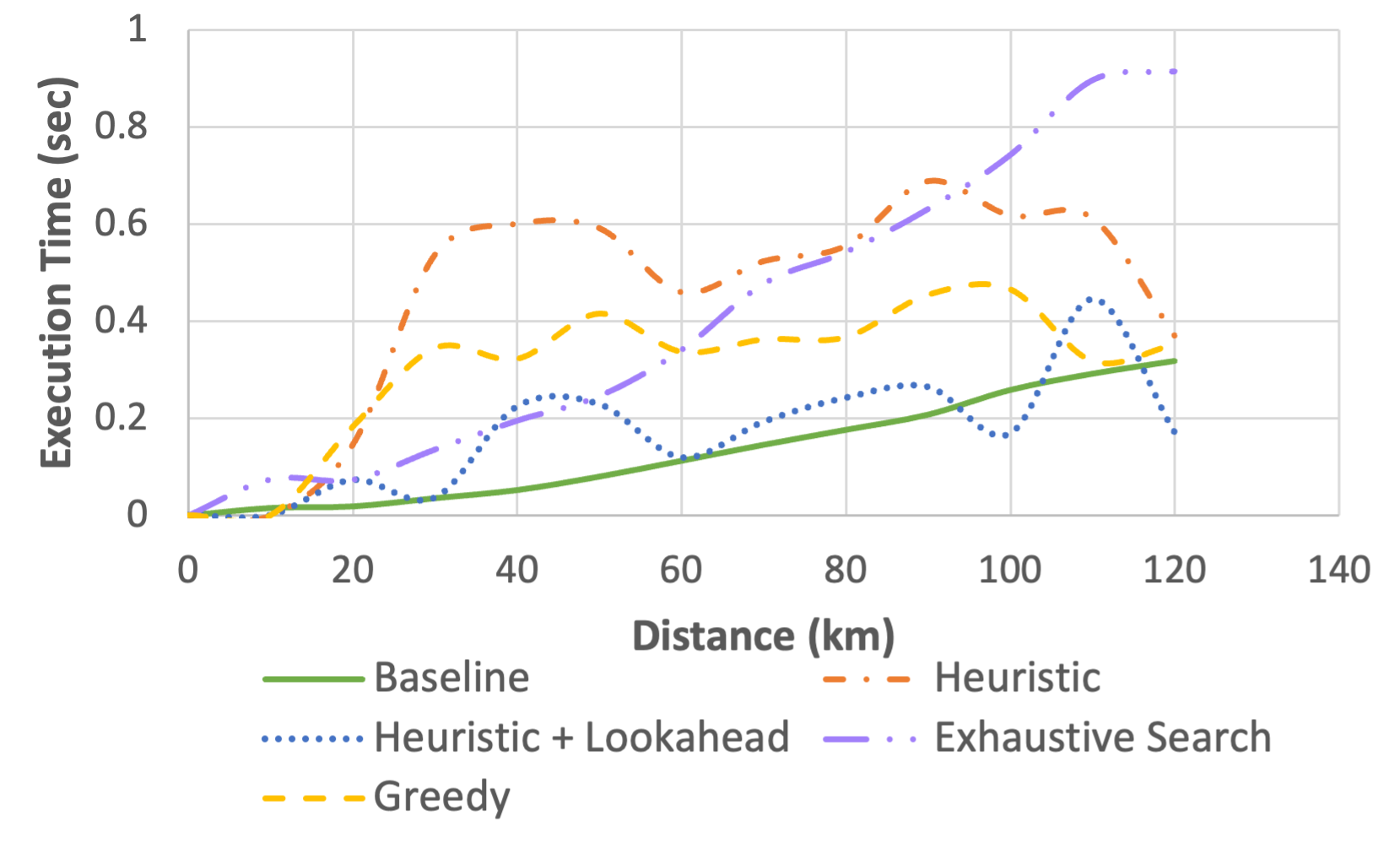}
  \captionof{figure}{Average Execution Times}
  \label{fig:executiontime}
\end{figure}

In the fourth experiment, we evaluate the packages delivery time performance for successful requests only. As shown in Fig. \ref{fig:deliverytime}, the proposed heuristic and the optimization with lookaheads outperforms the greedy algorithm which follows a safe but slow approach. This proves that changing speed during the composition has a great impact on the travel times and node times that compose the delivery time. The lookaheads optimization improves the heuristic since it is able to skip a charging stop when possible. Because the greedy approach is very slow, it may cause delivery failures due to late packages arrival as it is very far from the no failure baseline time. The exhaustive search approach as expected outperforms the rest but comes with the cost of high execution times which would have been exponential if not capped by the top 100 paths as shown in Fig. \ref{fig:executiontime}. The execution time of the lookaheads optimization is less than the greedy and the heuristic since less node stops means less assessments of speed. Moreover, there are less nodes to consider as there are more chances to find better nodes with shorter paths to the destination.


\section{Conclusion}
For swarm-based drone delivery services, we presented a novel failure-sentient framework. The framework ensures that soft failing drones in a swarm do not impede the successful package delivery to a consumer. To predict the time of failures and the failure uptime, the framework includes a weighted continuous federated learning prediction module. Subsequently, we conducted an evaluation to determine the severity of the failure at both the drone and swarm levels. The results obtained were then utilized to make informed decisions regarding the optimal path within the composition module.  Using a speed-based heuristic approach with lookaheads optimization, the predictions are used to compose an optimal set of services. The experimental results demonstrated the efficacy of our proposed method in terms of prediction accuracy, delivery times, and execution times. In terms of delivery times, the proposed composition approach outperformed the greedy algorithm. The optimisation of lookaheads improves the delivery times of our proposed heuristic.

\section*{Acknowledgment}
This research was partly made possible by DP220101823 and LE220100078 grants from the Australian Research Council. The statements made herein are solely the responsibility of the authors.

\bibliographystyle{ieeetr}
\bibliography{scholar}

\end{document}